\documentclass[conference]{IEEEtran}
\IEEEoverridecommandlockouts

\usepackage{cite}
\usepackage{amsmath,amssymb,amsfonts}
\usepackage{algorithmic}
\usepackage{graphicx}
\usepackage{textcomp}
\usepackage{xcolor}
\usepackage{multirow}
\usepackage{tikz}
\usepackage{comment}
\usepackage{svg}
\usepackage{url}
\usepackage{hyperref}
\usetikzlibrary{mindmap,trees,arrows.meta,positioning}
\usetikzlibrary{shadows,arrows}

\usepackage{booktabs,colortbl}

\def\BibTeX{{\rm B\kern-.05em{\sc i\kern-.025em b}\kern-.08em
    T\kern-.1667em\lower.7ex\hbox{E}\kern-.125emX}}

\newcommand\copyrighttext{%
  \footnotesize This work has been submitted to the IEEE for possible publication. Copyright may be transferred without notice, after which this version may no longer be accessible.}
\newcommand\copyrightnotice{%
\begin{tikzpicture}[remember picture,overlay]
\node[anchor=south,yshift=10pt] at (current page.south) {\fbox{\parbox{\dimexpr\textwidth-\fboxsep-\fboxrule\relax}{\copyrighttext}}};
\end{tikzpicture}%
}
    
\begin{document}

\newcommand{\ACRONYM}{CRADLE}

% RPTU colors
\definecolor{rptured}      {HTML}{e31b4c}
\definecolor{rptulblue}    {HTML}{6ab2e7}
\definecolor{rptudblue}    {HTML}{042c58}
\definecolor{rptulgreen}   {HTML}{26d07c}
\definecolor{rptudgreen}   {HTML}{006b6b}
\definecolor{rptuorange}   {HTML}{ffa252}
\definecolor{rptupurple}   {HTML}{4c3575}
\definecolor{rptupink}     {HTML}{d13896}
\definecolor{rptubluegray} {HTML}{507289}
\definecolor{rptugreengray}{HTML}{77b6ba}

\title{CRADLE: Conversational RTL Design Space Exploration with LLM-based Multi-Agent Systems
\thanks{This paper was partly funded by the German Federal Ministry of Research, Technology and Space as part of the ``SUSTAINET-GuarDian'' and ``Chipdesign Germany'' projects under grant numbers 16KIS2248 and 16ME0890.}
}

\author{\IEEEauthorblockN{Lukas Krupp, Maximilian Sch\"offel, Elias Biehl, and Norbert Wehn}
\IEEEauthorblockA{RPTU University of Kaiserslautern-Landau, Kaiserslautern, Germany}
}

\maketitle
\copyrightnotice

\begin{abstract}
This paper presents \ACRONYM{}, a conversational framework for design space exploration of RTL designs using LLM-based multi-agent systems. Unlike existing rigid approaches, \ACRONYM{} enables user-guided flows with internal self-verification, correction, and optimization. We demonstrate the framework with a generator-critic agent system targeting FPGA resource minimization using state-of-the-art LLMs. Experimental results on the RTLLM benchmark show that \ACRONYM{} achieves significant reductions in resource usage with averages of 48\% and 40\% in LUTs and FFs across all benchmark designs.
\end{abstract}


\section{Introduction}
The increasing complexity of state-of-the-art system-on-chips continuously drives the need for improved design productivity. As a result, tools and methods such as high-level synthesis (HLS) for dataflow-dominant applications have been introduced to accelerate development cycles. Nevertheless, most digital designs are still specified using hardware description languages (HDLs) like Verilog or VHDL. HDLs allow for fine-grained control over critical physical implementation parameters such as area and timing. However, the low abstraction level of HDLs expands the design space and makes the register-transfer level (RTL) design process both time-consuming and error-prone. 

To address these productivity bottlenecks, artificial intelligence (AI) methods are being increasingly integrated into electronic design automation (EDA) workflows~\cite{b1}. Large language models (LLMs) have emerged as promising tools for a range of RTL design tasks, including code generation, optimization, verification, and comprehension. Recent research shows that LLM-based agent systems outperform standalone LLMs on such tasks by strategic planning, orchestrating EDA tools, and maintaining context~\cite{b2}. These systems perform self-verification and self-correction of HDL code via integration with RTL simulation tools~\cite{b3, b4, b5}, and automate design space exploration (DSE) using synthesis and backend tools~\cite{b7, b8}.

However, existing solutions overlook the interactive and conversational capabilities of LLM-based agents. Current approaches use rigid input-output structures, limiting the designer's ability to guide exploration or incorporate domain expertise. This lack of interactivity is especially problematic in optimizing complex legacy designs, where extensive prior knowledge and established hierarchical code bases exist.

In this work, we present the following key contributions:
\begin{itemize}
    \item We propose \ACRONYM{}, a novel framework for conversational DSE via LLM-based agents with internal self-verification, self-correction, and self-optimization.
    \item We demonstrate the framework in a case study focused on minimizing FPGA resource utilization of RTL designs.
    \item We provide an evaluation on the RTLLM benchmark, utilizing state-of-the-art LLMs (GPT-4.1 and o4-mini).
\end{itemize}

\section{Proposed Framework}
\begin{figure}[htbp]
\centerline{\includegraphics[width=0.5\textwidth]{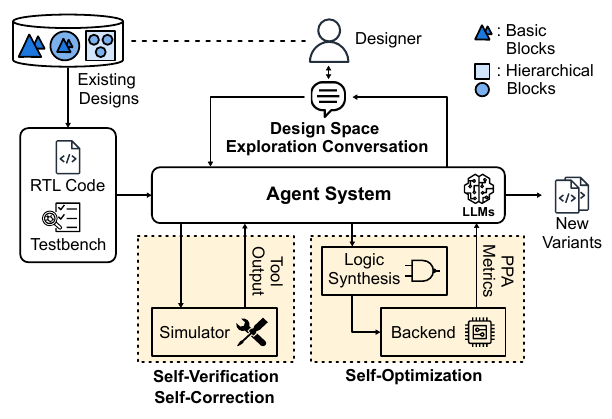}}
\caption{Overview of the \ACRONYM{} framework.}
\label{fig:framework}
\end{figure}

The proposed framework illustrated in Fig.~\ref{fig:framework} enables interactive DSE for hierarchical RTL designs through conversations with a multi-agent system. An example for such a design is a fast Fourier transform (FFT) accelerator. The top-level module might be a fully-parallel FFT architecture (square). This block contains submodules (circle) such as radix-2 or radix-4 butterfly units, which are further decomposed into basic arithmetic blocks (triangle) like adders and multipliers.

The agent system has access to the RTL code and testbenches of existing designs. Designers may initiate high-level optimization goals, such as generating variants of existing blocks optimized for power, performance, or area (PPA). Alternatively, they can inject domain expertise by instructing the agent to apply specific architectural choices like converting a fully-parallel FFT into a sequential pipelined version.

\begin{figure*}[t!]
  \centering
  \includegraphics[width=\linewidth]{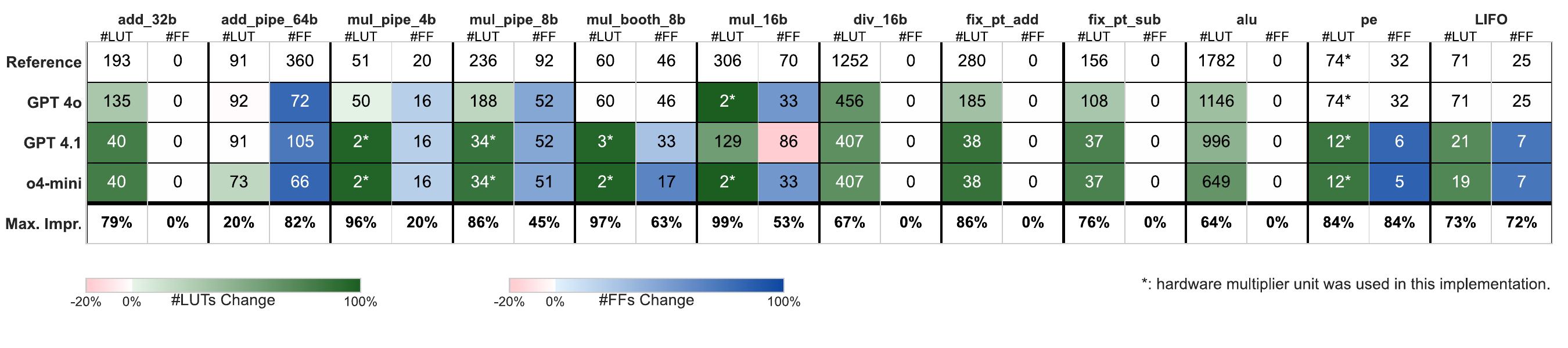}
  \caption{Resource usage (LUTs, FFs) after place-and-route on ECP5 for RTLLM designs using \ACRONYM{} with different LLMs vs. the reference implementation.}
  \label{fig:results}
\end{figure*}

The agent system is configurable, with an arbitrary number of agents and diverse coordination patterns. Agents can employ different LLMs and strategies, enabling the exploration of agent system setups. The agent system has access to an RTL simulator to facilitate self-verification (syntax checking, testbench execution) and self-correction. Furthermore, it uses synthesis and backend tools for PPA-focused self-optimization.

\section{Experimental Setup}
To assess the effectiveness of \ACRONYM{}, we conducted a case study focused on minimizing resource utilization of designs implemented on an FPGA. The Lattice ECP5 FPGA was selected as the target platform due to its industrial relevance. The reference designs and testbenches from the RTLLM benchmark~\cite{b10} served as inputs to the agent system. We implemented the framework in Python based on the Google Agent Development Kit (ADK) and instantiated an agent system adopting the generator-critic pattern, commonly used for code generation and optimization~\cite{b9}. 

An optimizer agent reviews the initial RTL design and, informed by the user-specified objective, formulates an optimization plan. The agent interfaces with the open-source tools \textit{Yosys} for synthesis and \textit{nextpnr} for place-and-route to analyze the cell usage of the current RTL code. A rewriter agent executes the optimization plan by modifying the code. This agent has access to \textit{ModelSim} and the testbench to ensure syntactic and functional correctness of the rewritten variants. Following each round of rewriting, the system revisits the code and optimization plan to further improve the resource usage. 

The agent system performs up to three internal iterations before returning the best result to the designer.
Experimental evaluation is conducted using the state-of-the-art LLMs o4-mini for reasoning and GPT-4.1 for completion tasks by OpenAI. GPT-4o is employed as an additional reference.

\section{Results}
We compared the cell utilization of the reference benchmark implementations with the best agent-generated variants. Across the 50 designs in the RTLLM benchmark, the number of resource-optimized variants varied with the underlying LLM. o4-mini achieved improvements in 41 designs, compared to 37 with GPT-4.1 and 27 with GPT-4o. On average, o4-mini reduced lookup table (LUT) usage by 48\% and flip-flop (FF) usage by 40\% across all designs, while GPT-4.1 achieved reductions of 33\% and 27\%, and GPT-4o yielded reductions of 23\% and 14\%, respectively. Fig.~\ref{fig:results} presents the detailed results for the 10 best designs from o4-mini and GPT-4.1. There is a substantial overlap among these top designs, resulting in 12 unique cases.

The o4-mini agents consistently delivered the largest improvements, for all designs exceeding or matching the results of the other models.  The most significant reduction in LUTs was observed for the \texttt{mul\_16b} module, while the \texttt{pe} module demonstrated the highest reduction in FFs. These results illustrate the effectiveness of reasoning models, particularly within a generator-critic pattern, for optimization tasks demanding analysis and planning. The demonstrated reductions in resource usage validate the capability of the conversational agent system to follow user guidance and leverage EDA tools for self-verification, self-correction, and self-optimization.

\section{Conclusion}
In this paper, we introduced a conversational framework for interactive DSE of RTL designs with LLM-based agents. Our approach enables automated self-verification, correction, and PPA-focused optimization, while allowing user guidance throughout the process. Experimental results show significant improvements in resource usage of FPGA designs, with average reductions in LUTs and FFs of 48\% and 40\% across the RTLLM benchmark. Future work will focus on the human-in-the-loop optimization and industry-relevant IP cores.

\end{document}